\documentclass[conference]{IEEEtran}
\IEEEoverridecommandlockouts

\usepackage{cite}
\usepackage{amsmath,amssymb,amsfonts}
\usepackage{algorithmic}
\usepackage{graphicx}
\usepackage{textcomp}
\usepackage[dvipsnames]{xcolor}
\usepackage{caption}
\usepackage{subcaption}
\usepackage{tabularx}
\usepackage{booktabs}
\usepackage{multirow}
\usepackage{collcell}
\usepackage{hhline}
\usepackage{textcomp}

\usepackage{graphicx}
\usepackage{tikz}
\usepackage{forest}
\usetikzlibrary{trees,positioning,shapes,shadows,arrows.meta}

\def\BibTeX{{\rm B\kern-.05em{\sc i\kern-.025em b}\kern-.08em
    T\kern-.1667em\lower.7ex\hbox{E}\kern-.125emX}}

\begin{document}

\title{Exploring the Adversarial Robustness of CLIP for AI-generated Image Detection}

\author{
\IEEEauthorblockN{Vincenzo De Rosa, Fabrizio Guillaro, Giovanni Poggi, Davide Cozzolino and Luisa Verdoliva} 
\IEEEauthorblockA{University Federico II of Naples, Italy \\
Email: \{vincenzo.derosa3, fabrizio.guillaro, poggi, davide.cozzolino, verdoliv\}@unina.it}
}
\maketitle

\begin{abstract}
In recent years, many forensic detectors have been proposed to detect AI-generated images and prevent their use for malicious purposes.
Convolutional neural networks (CNNs) have long been the dominant architecture in this field and have been the subject of intense study.
However, recently proposed Transformer-based detectors have been shown to match or even outperform CNN-based detectors, especially in terms of generalization.
In this paper, we study the adversarial robustness of AI-generated image detectors, focusing on Contrastive Language-Image Pretraining (CLIP)-based methods that rely on Visual Transformer (ViT) backbones and comparing their performance with CNN-based methods.
We study the robustness to different adversarial attacks under a variety of conditions and analyze both numerical results and frequency-domain patterns.
CLIP-based detectors are found to be vulnerable to white-box attacks just like CNN-based detectors. However, attacks do not easily transfer between CNN-based and CLIP-based methods.
This is also confirmed by the different distribution of the adversarial noise patterns in the frequency domain.
Overall, this analysis provides new insights into the properties of forensic detectors that can help to develop more effective strategies.
\end{abstract}


\section{Introduction}
AI-generated images are here to stay.
People no longer pay much attention to the nature of an image, whether it is generated by a conventional device, created by a neural network or, more often, acquired by a smartphone with plenty of AI-based enhancement filters.
However, there are many situations, in journalism, politics, or the judiciary, where establishing the nature of an image, real or synthetic, is of great importance.
In recent years there has been intense research on this topic.
A number of CNN-based detectors have been proposed
which can easily recognize a synthetic image when it is generated by an AI model seen in the training phase \cite{lin2024detecting}.
Unfortunately, their performance degrades sharply on images generated by new models (not an uncommon case) or impaired by compression or resizing \cite{tariang2024synthetic}.
Very recently, however, several new detectors have been proposed based on transformer architectures or that rely on features extracted from CLIP \cite{ojha2023towards, sha2023defake, cozzolino2024raising}.
These approaches show very promising result in terms of generalization to unseen synthetic samples. This is in-line with current findings in computer vision that state that the transformer’s self-attention-like architecture is a key ingredient for improving the performance on out-of-distribution samples \cite{bai2021transformer}.

In this work we want to investigate the robustness to adversarial attacks of such powerful forensic detectors and compare them with CNNs.
Adversarial deep learning is a thriving field of research in its own right.
This is even more true for forensic detectors \cite{barni2018adversarial}.
In fact, to distinguish real images from those generated by AI they rely on subtle and easily perturbed traces,
a sort of artificial fingerprints that characterize generative architectures \cite{marra2019DoGAN, corvi2023intriguing}.
Early works have demonstrated that deepfake detectors can be successfully attacked
in both a white-box and black-box scenario \cite{carlini2020evading}
and that attacks prove robust to compression codecs \cite{hussain2021adversarial},
making them a very concrete threat.

\begin{figure}
    \centering
    \includegraphics[width=0.99\linewidth, page=2,clip, trim=70 0 70 0]{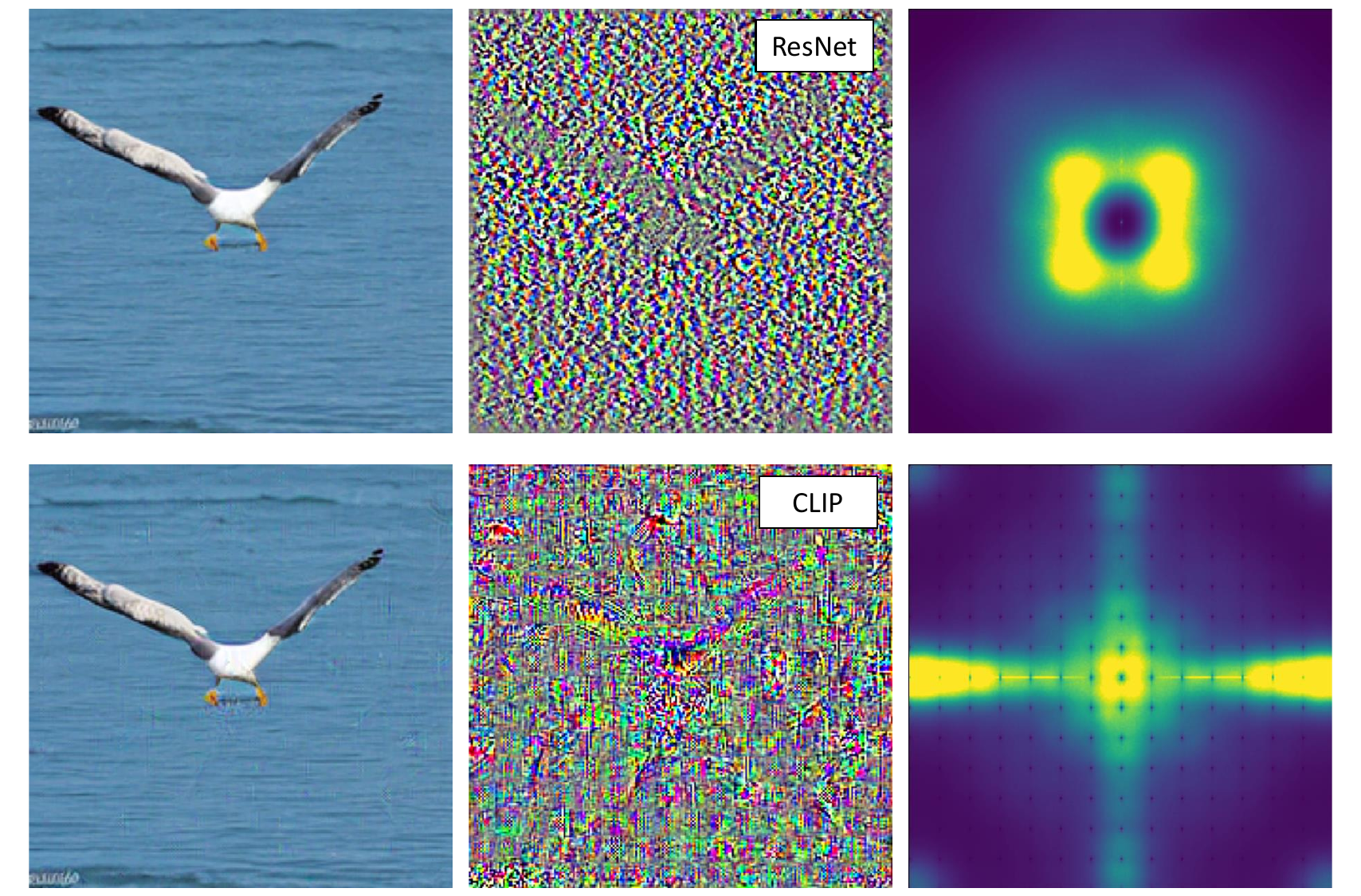}
    \caption{{\bf CLIP vs. ResNet.}
    $l_2$-PGD attacks to ResNet-based (top) and CLIP-based (bottom) detectors.
    From left to right, attacked image, magnified adversarial perturbation, average spectrum of the adversarial noise. 
    Attacks to Transformer-based detectors work on lower frequencies than attacks to CNN-based detectors do. 
    In addition they present a clear cross-shaped directional spectrum, as also shown in \cite{bhojanapalli2021understanding} for image classification, 
    which is due to the patch-wise processing before the transformer blocks.}
    \label{fig:teaser}
\end{figure}

\begin{figure*}
    \centering
    \includegraphics[width=0.99\linewidth,page=3,clip,trim=0 340 0 0]{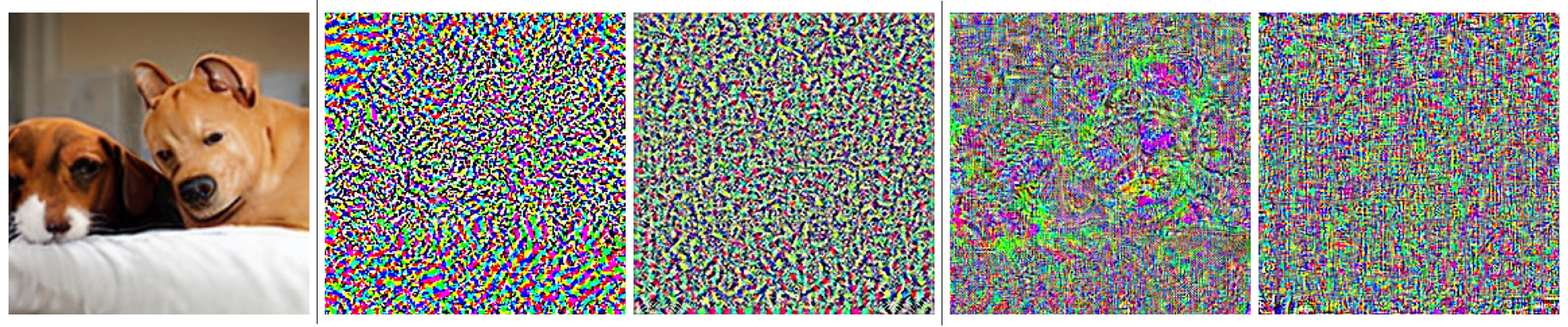} 
    \caption{From left to right: attacked image; magnified noise patterns generated by RWA and UA for a CNN-based detector; magnified noise patterns generated by RWA and UA for a CLIP-based detector. Even for these attacks we can make similar observations as done for $l_2$-PGD in Fig. 1: CLIP-based attacks are more structured and show clear regular patterns.}
    \label{fig:att_noise}
\end{figure*}

However, these works limit their analyses to CNN-based detectors neglecting architectures based on transformers. Here we want to fill this gap and study the behavior of CLIP-based forensic detectors in the presence of adversarial attacks and compare with CNN-based detectors. It is worth to note that in computer vision there has been signficant attention to analyze the adversarial robustness of ViT compared to CNN-based solutions for the generic task of image classification \cite{bhojanapalli2021understanding, bai2021transformer, benz2021adversarial, mahmood2021robustness}, while there is a lack of such analysis in the forensics field.
In our study we will give a special attention to the key issue of attack transferability.
Indeed, fooling a known detector is little more than a classroom exercise: the real goal is to fool all (or most) of them.
In addition to presenting numerical results,
we will conduct a careful analysis of the adversarial noise patterns that emerge in this process,
both in the spatial and frequency domains (Fig.~\ref{fig:teaser} and Fig.~\ref{fig:att_noise}).
The results will provide interesting clues about how these detectors work
and suggest possible research directions for their further improvement in terms of performance or robustness.

\section{Preliminaries}

\subsection{Forensic detectors}

In our experiments, we consider four CNN-based detectors and four CLIP/ViT-based detectors, summarized in Table~\ref{tab:detectors}.
In the CNN-based detector family, 
we first consider the popular detector proposed in \cite{wang2020cnn}, a simple ResNet50 pre-trained on ImageNet and optimized for synthetic image detection with compression-based augmentation and blurring.
The second detector \cite{gragnaniello2021GAN} differs only for one architectural change, 
the removal of subsampling in the first layer of the network. 
The same architecture is used in the third detector \cite{corvi2023detection} 
which adopts stronger live data augmentation to improve robustness, 
including blurring, scaling, CutOut, compression, noise addition, and color jittering.
For the last CNN-based detector, we consider ConvNext Tiny \cite{liu2022convnet}, comparable to ResNet50 in number of parameters, also modified to avoid downsampling in the first layer.

The first two transformer-based detectors implement the strategy proposed in \cite{ojha2023towards}, 
where features are extracted from a large pre-trained model, CLIP/ViT-L \cite{radford2021learning} in the first case, and the larger EVA-CLIP/ViT-g \cite{sun2023eva} in the second case, 
followed by a single-layer neural network fine-tuned for synthetic image detection.
The last two detectors adopt an architecture consisting of CLIP/ViT-B followed by two linear layers 
and the whole network is fine-tuned for synthetic image detection using two different augmentation strategies 
proposed in \cite{wang2020cnn} and \cite{corvi2023detection}, respectively.
In Table~\ref{tab:detectors}, we summarize the main features of such detectors.

\begin{table}
    \centering
    \setlength{\tabcolsep}{2pt}
    \resizebox{0.9\linewidth}{!}
    {\begin{tabular}{cclcc} \toprule
       & Family & Network & Fine-tuning & Aug. \\  
       \cmidrule(r){2-2} \cmidrule(r){3-3} \cmidrule(lr){4-4} \cmidrule(r){5-5}
       (1) & CNN & ResNet50 & e2e & $\ast$ \\
       (2) & CNN & ResNet50\dag & e2e & $\ast$ \\
       (3) & CNN & ResNet50\dag & e2e & $\ast\ast$ \\
       (4) & CNN & ConvNeXt Tiny\dag & e2e & $\ast\ast$ \\ 
       (5) & CLIP & ViT-L + 1 FC layer  & F & $\ast$ \\ 
       (6) & CLIP & ViT-g + 1 FC layer  & F & $\ast$ \\ 
       (7) & CLIP & ViT-B + 2 FC layers & e2e & $\ast$ \\ 
       (8) & CLIP & ViT-B + 2 FC layers & e2e & $\ast\ast$ \\
       \bottomrule
    \end{tabular}}
    \\
    \resizebox{0.9\linewidth}{!}{
    \begin{tabular}{l}
    ($\ast$)~ compression and blurring \\
    ($\ast\ast$)~ compression, blurring, scaling, cut-out, noise addition, jittering \\    
    \end{tabular}
    }
    \caption{List of detectors. All methods were trained using the dataset proposed in \cite{corvi2023detection} that includes COCO/LSUN and latent diffusion. ViT-based networks are followed by 1 or 2 fully connected (FC) layers. Models with \dag~were modified to not perform downsampling in the first layer. Fine-tuning is performed end-to-end on the entire network (e2e) or only on the last linear layer, keeping the backbone frozen (F).}
    \label{tab:detectors}
\end{table}

\begin{figure*}
    \centering
    \includegraphics[width=0.83\linewidth,page=1,clip,trim=35 60 40 0]{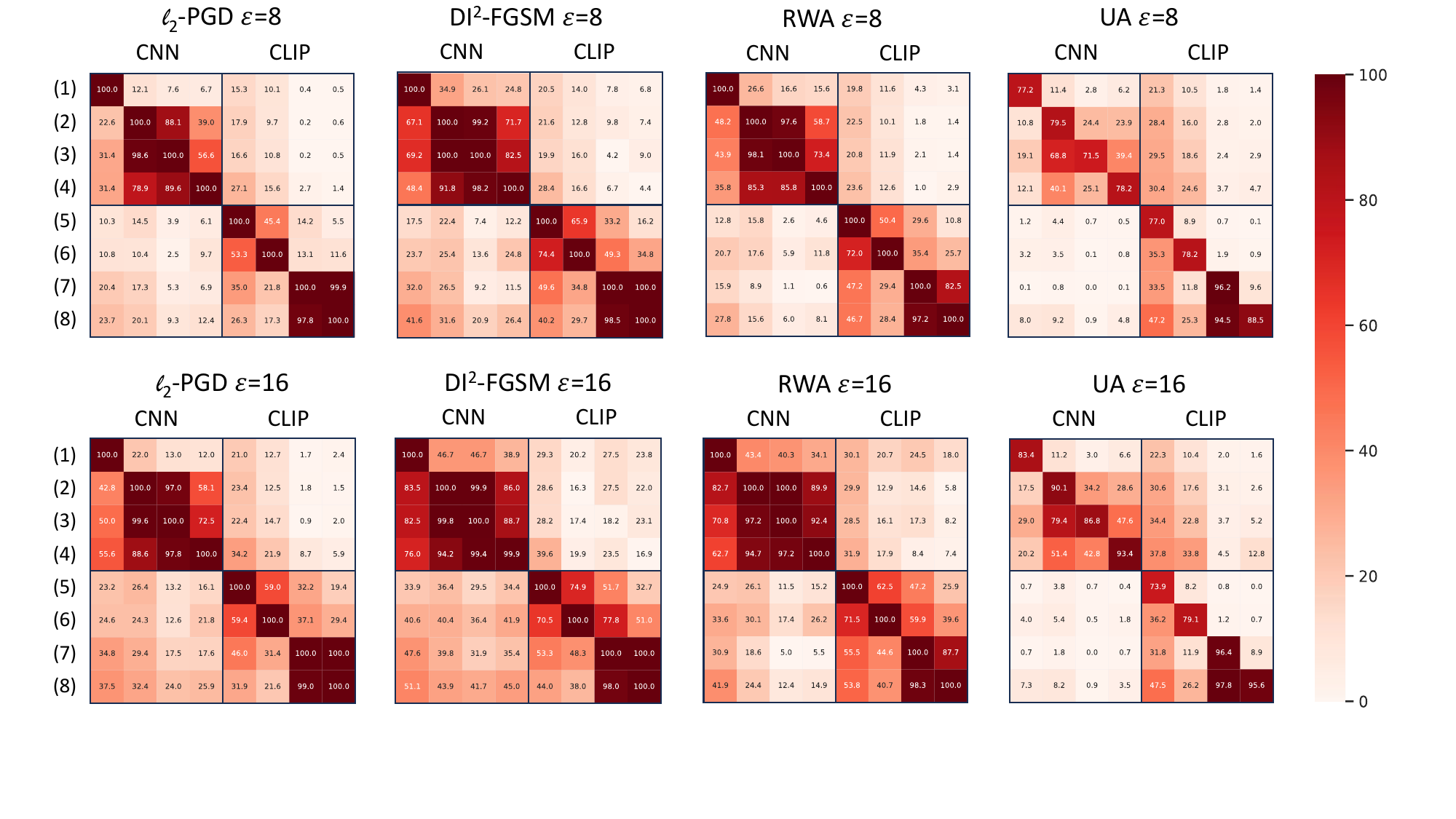}
    \caption{Successful Attack Rate of four attacks (PGD, DI$^2$-FGSM, RWA, UA) at two strength levels ($\epsilon$=8, $\epsilon$=16) on the eight detectors of Tab.~\ref{tab:detectors}.
    Cells on the diagonal correspond to white-box attacks. Off-diagonal cells correspond to transferred attacks.
    }
    \label{fig:cross}
\end{figure*}

\subsection{Adversarial attacks}

Neural networks are known to be vulnerable to adversarial attacks.
By adding subtle adversarial noise to the input image, it is possible to fool the target classifier into predicting a wrong label.
Forensic detectors are no exception as shown in the literature for applications as diverse as image forgery detection and source attribution \cite{barni2022adversarial}.
Similarly, several papers \cite{carlini2020evading, neekhara2021adversarial, jia2022exploring}
have shown that deepfake detectors can be easily attacked in a white-box scenario, i.e., when all the details of the detector are perfectly known.
Even if the attacker has only partial knowledge of the detector, an attack designed for a surrogate model can be used with good success rates \cite{carlini2020evading, hussain2021adversarial}.
Conversely, achieving attack transferability in a zero-knowledge scenario is non-trivial, 
especially if the surrogate and target architectures are significantly different or trained with very different protocols \cite{zhao2022making, wang2021perception}.

Here, we focus on white-box attacks
since our aim is to investigate the fundamental properties of different families of detectors in fully controlled conditions.
However, we will also investigate in depth attack transferability, as it pertains the scenario of highest practical interest.

When the detector is perfectly known together with its parameters, $\theta$, the attacker can compute all possible gradients of interest.
This is exploited in the Fast Gradient Sign Method (FGSM) proposed originally in \cite{goodfellow2015explaining}.
By computing the gradient of the model loss $L(\theta,x,y)$ with respect to the input image $x$
one finds the direction that maximizes the disruptive effect of the attack for a given input perturbation.
Following this idea, in FGSM the adversarial sample is computed as
\[
    x_a = x + \epsilon\, {\rm sign}[\nabla_x L(\theta,x,y)]
\]
Note that,
to enforce a strict bound on the image distortion,
only the sign of the gradient is taken, with the parameter $\epsilon$ controlling the strength of the perturbation.\footnote{The same parameter $\epsilon$ is used also in other attacks to control the distortion introduced in the image and is commonly taken to indicate the attack strength.}
In this work we will consider four different attacks, briefly described below.

The {\bf Projected Gradient Descent (PGD)} \cite{madry2018towards} is an iterative attack that has been demonstrated to be highly effective and sophisticated in comparison to FGSM.
Indeed, the single iteration of FGSM is optimal only when $\epsilon \to 0$, while it may be largely sub-optimal for stronger attacks.
So, PGD builds the adversarial example starting from $x^0=x$ and updating it iteratively as $x^{t+1} = \Pi_{B(x)} \left\{x^t + \alpha\,\, {\rm\bf N}[\nabla_{x^t} L(\theta,x^t,y)] \right\}$
where $\alpha \ll 1$, ${\rm\bf N}[\cdot]$ indicates normalization with reference to the adopted norm, and $\Pi_{B(x)}$ projects the result on a ball defined by the norm, centered on $x$, and with radius $\epsilon$. 
PGD is much more effective than FGSM but still computationally affordable.
Moreover, in \cite{madry2018towards} it is claimed to ensure good transferability to other architectures. In any case, it is a de-facto baseline for all studies on adversarial attacks.

The {\bf Diverse Inputs I-FGSM (DI$^2$-FGSM)} \cite{xie2019improving} incorporates the input diversity strategy with an iterative version of FGSM (I-FGSM) attack to improve transferability. 
It applies transformations to the inputs with a certain probability at each iteration. This approach mitigates the overfitting by diversifying the input data. The image transformations employed are random resizing and random zero padding.

The {\bf Robust White-Box Attack (RWA)} \cite{hussain2021adversarial} is structurally similar to PGD
with some important changes meant to address forensic applications.
First of all, it is explicitly developed for deepfake detection, hence a native two-class problem.
More important, it takes into account the fact that, in this field,
images are usually subject to various kinds of impairing transforms (compression, resizing, blurring) before being analyzed by the detector.
Therefore, a generalized loss function is defined (computed in practice by sample averaging) which includes transformed versions of the original image to gain robustness to this adverse scenario.

Finally, we include the {\bf Universal Attack (UA)} proposed in \cite{moosavi2017universal}. Here, the goal is not to attack a given image
but rather to design a single adversarial noise sample, say $\Delta x$,
such that the classifiers chooses a wrong label, $y(x+\Delta x) \neq y(x)$, for most input images. The desired perturbation is obtained by accumulating the gradients computed on a large sample of input images. Experiments show UA to be surprisingly effective, although less than targeted attacks. We include it in our analysis because it fits the classifier as a whole, hence we hope that it helps shading light on specific properties of the detectors under investigations.

\begin{figure}
    \centering
    \includegraphics[width=0.99\linewidth,page=1,clip,trim=0 0 0 0]{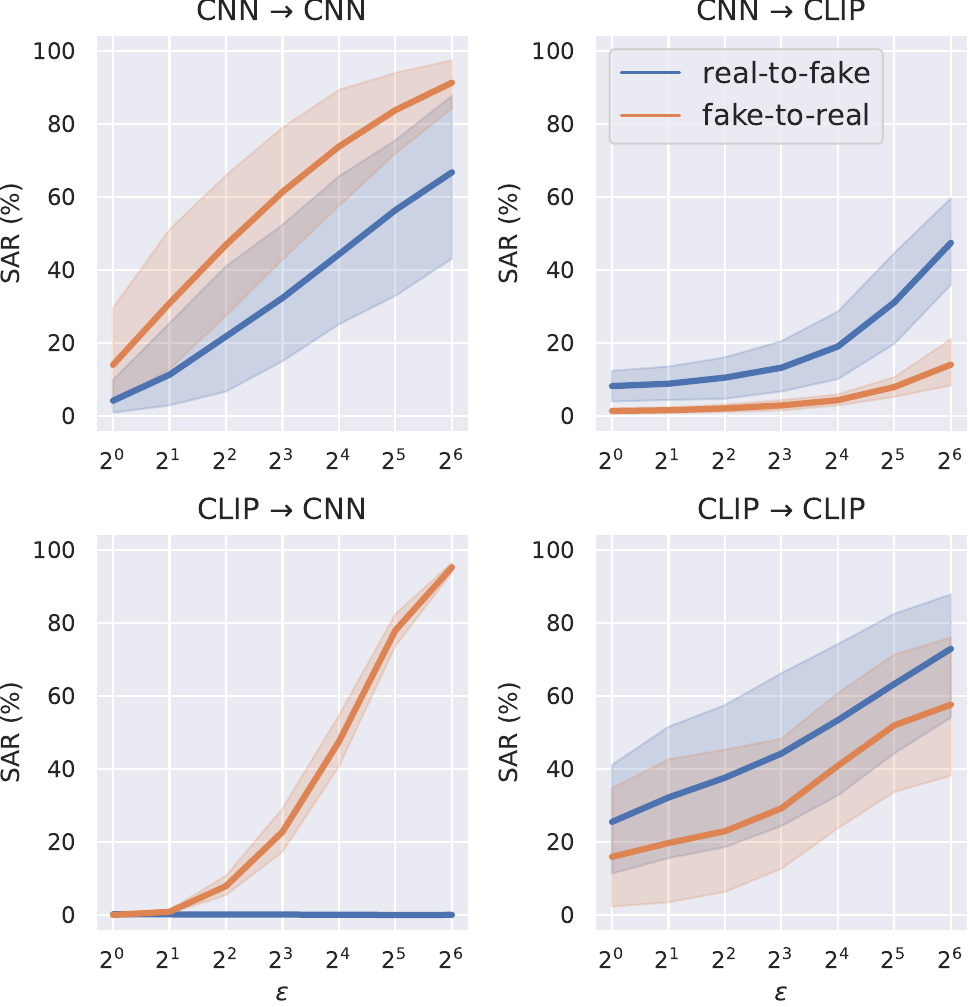}
    \caption{Transferability of $l_2$-PGD attacks as a function of the attack strength $\epsilon$. 
    Solid lines represent the average SAR and colored bands its standard deviation. 
    We consider target and source detectors belonging to the same family (top-left, bottom-right) or different families (top-right, bottom-left).}
    \label{fig:intra_inter_class}
\end{figure}

\section{Attack transferability}

The attacks are evaluated in terms of Success Attack Rate (SAR) 
measured on 1000 real images of the COCO dataset and 1000 images generated by Latent Diffusion Models (LDMs). 
To ensure compatibility with all detectors, images are central cropped to $224\times 224$ pixels.
In Fig.~\ref{fig:cross} we report synthetic numerical results extracted from our experimental campaign.
The figure displays eight matrices,
corresponding from left to right to the four attacks (PGD, DI$^2$-FGSM, RWA, UA)
and from top to bottom to two attack strengths
($\epsilon$=8 lighter, and $\epsilon$=16 stronger), corresponding to an average PSNR larger than 36dB and 30dB, respectively, which guarantee to avoid visual artifacts.
In each matrix, the cell $(i,j)$ reports the average success rate
observed on the $j$-th detector (target) using adversarial samples designed on the $i$-th detector (source),
where $i,j \in \{1,\ldots,8\}$ span the eight detectors listed in Tab.~\ref{tab:detectors}, 4 CNN-based and 4 CLIP-based.
Cells on the diagonal show the white-box SAR for each detector.
The image-targeted attacks, PGD, DI$^2$-FGSM and RWA, have a uniform success rate of 100\% on all detectors and both strengths.
The universal attack is somewhat less effective but SAR is always over 70\%.

More interesting are the results on attack transferability, given by the off-diagonal cells. As expected, DI$^2$-FGSM attack exhibits superior transferability in comparison to other attacks.
We now focus on the upper-left matrix, corresponding to PGD@$\epsilon$=8, since the other matrices show only minor differences.
The color coding of the cells (darker as the attack gets more successful) allows one to catch the big picture at a glance.
Contrary to some claims in the literature, attacks are not easily transferable.
To be more precise, SAR is always very low whenever source and target models belong to different families, CNN vs. CLIP.
This makes perfect sense, considering the profound architectural difference that characterize these models.
However, mixed results are observed also within the same family.
For example, attacks seems to transfer very well between detectors (2), (3) and (4), but not from this group to detector (1).
This is very interesting, considering that detector 4 uses a ConvNeXt backbone while all the others use ResNet50, suggesting that the backbone does not impact as much on the detector behavior as other architectural choices, like removing subsampling from the first layer.
In hindsight, this analysis provides valuable information on what counts for detection and what is irrelevant.
Similar results (not reported here) are observed with ProGAN images.
The non-transferability of the attack between different families is confirmed by a SAR less than 46\% for $\epsilon$=8 and 64\% for $\epsilon$=16.

With the help of Fig.~\ref{fig:intra_inter_class} we better investigate attack transferability,
analyzing for $l_2$-PGD separately attacks to real and synthetic images and considering a wide range of attack strengths, from $\epsilon$=1 to $\epsilon$=64.
Each chart reports the average SAR obtained with source detector in one family, CNN or CLIP, and target detector in another family.
Of course, even when the two families coincide, only different source and target detectors are considered.
The top-left and bottom-right charts are for same-family cases, CNN$\to$CNN and CLIP$\to$CLIP.
Results are as expected, with a pretty good transferability that increases smoothly with attack strength.
The other two charts are more interesting.
In particular, they show that CNN detectors are vulnerable to fake-to-real attacks, the most dangerous for real-world cases.
On the contrary, CLIP detectors are weaker with respect to real-to-fake attacks, 
but only at high strengths, when image distortion cannot be neglected anymore.
This behavior will be more easily explained using the Fourier-domain analyses of next Section.

\begin{figure*}
    \centering
    \includegraphics{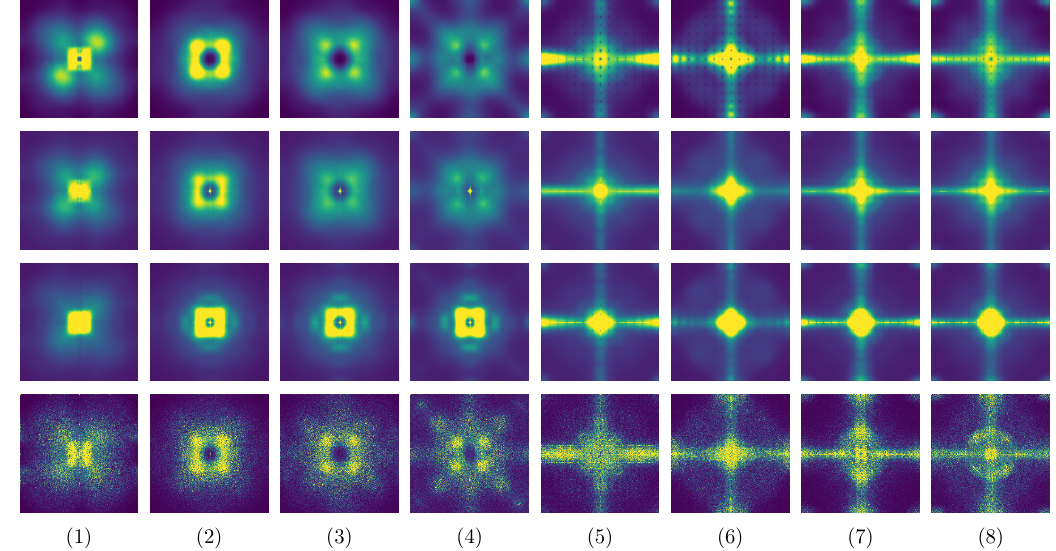}
    \caption{Power spectra of adversarial noise patterns generated by a specific attack (rows) on a selected detector (columns). From top to bottom: $l_2$-PGD, DI$^2$-FGSM, RWA, UA attacks. From left to right: detectors (1) to (8) listed in Table \ref{tab:detectors}.} 
    \label{fig:att_fft}
\end{figure*}

\section{Frequency-domain analysis}

\subsection{Power spectra of adversarial noise patterns}

It is well known that image generators cannot perfectly mimic the frequency domain behavior of natural images.
By looking at the Fourier spectrum of adversarial noise, we can gain insight into how different detectors analyze the images \cite{bai2022improving}.
In Fig.~\ref{fig:att_fft},
for each attack we show the average power spectra of the adversarial noises generated to attack the detectors.
In more detail, we generate 1000 adversarial noise patterns for real images and 1000 for fake ones, compute the squared modulus of their Fourier transform and average them to obtain the desired power spectrum.

The analysis reveals several interesting facts, first of all the differences between CNN-based and CLIP-based detectors.
CNN spectra\footnote{Short for ``power spectra of adversarial patterns generated to attack CNN-based detectors''.} 
exhibit strong components at medium-high frequencies while CLIP spectra are concentrated at medium-low frequencies. 
This is clearly related to the backbones since, as already noted in the literature \cite{park2022how}, 
CNNs are better than Transformers in capturing fine image details (high frequencies) and, conversely, the latter are better equipped to see non-local or long-range dependencies (low frequencies).
In addition, CNN spectra are approximately isotropic, while CLIP spectra are markedly cross-shaped.
This is due to the block processing (blocks of 14$\times$14 or 16$\times$16 pixels) performed by ViT in the initial stages, 
as also confirmed by the sinc-like behavior of the horizontal and vertical cuts of the spectra, with regular zeros, reminiscent of the transformation of a spatial rectangular window.
These large differences help explain the different behavior of CNN-based and CLIP-based detectors.
For example, in \cite{cozzolino2024raising} it was observed that CLIP-based detectors are largely immune to image autoencoding-based recycling attacks, which are very effective with CNN-based detectors.
Considering that such attacks mainly target the high frequencies of the signal, this fact is no longer a mystery.

On the other hand, the specific properties of the individual detectors are also important, as are the specific characteristics of the individual attacks.
For example, the spectra of detector (1), in contrast to other CNN-based detectors, are more concentrated at low frequencies.
We explain this by the presence of early subsampling in the detector architecture, which significantly attenuates high frequencies from the image.
The augmentation protocol also plays a role.
Detectors trained with strong augmentation, such as (3) and (8), consider a wider range of frequencies than their weakly augmented versions, (2) and (7).
Finally, note that the spectra obtained for DI$^2$-FGSM and the RWA attack are also more concentrated at low frequencies.
In our interpretation, this is to withstand post-processing actions that tend to erase fine details.

\subsection{On the complementarity conjecture}

\begin{figure}  
    \centering
    \includegraphics[width=0.49\linewidth,page=1,clip,trim=0 0 0 0]{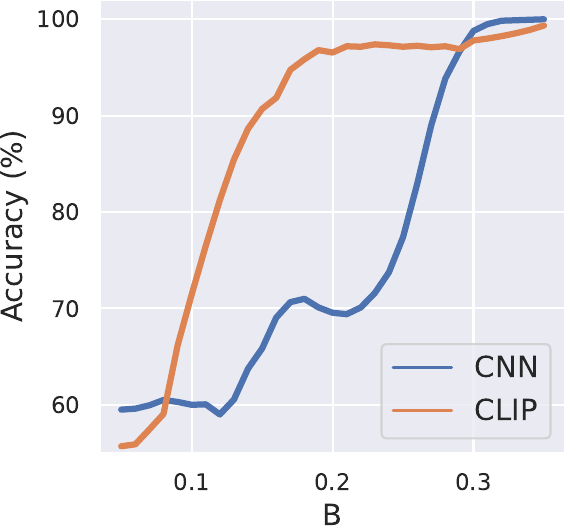}
    \includegraphics[width=0.49\linewidth,page=1,clip,trim=0 0 0 0]{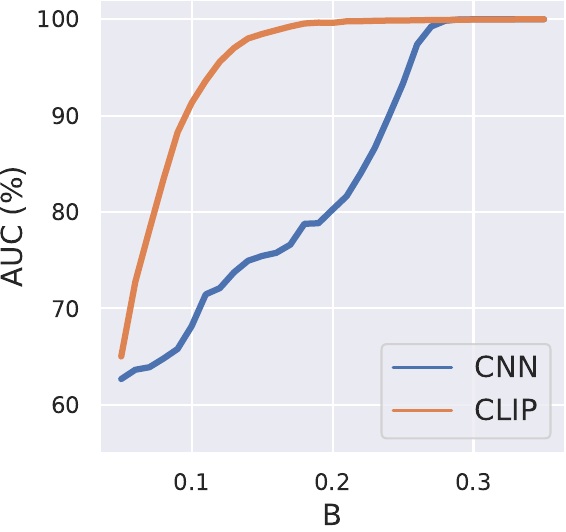}
    \caption{Performance of a CNN-based detector and a CLIP-based detector on low-pass filtered images as a function of the filter bandwidth $B$. Left: Accuracy, right: AUC.}    \label{fig:freq}
\end{figure}

Before concluding this work, we want to further elaborate on the most relevant concept that emerged from the previous analysis,
namely the apparent complementarity between CNN-based and CLIP-based detectors.
We have already hypothesized this fact in \cite{cozzolino2024raising}.
The analysis of adversarial noise patterns has further strengthened this hypothesis, 
which is also confirmed in \cite{park2022how}.
In Fig.~\ref{fig:freq}, we report the results of a further experiment along this line. 
The plots show the performance in terms of Accuracy (left) and Area under the ROC curve (right) of two comparable CNN-based (3) and CLIP-based (8) synthetic image detectors, 
as a function of the bandwidth $B$ of a low-pass filter used to remove high-frequency components.
The CLIP-based detector achieves its best performance at $B$=0.2, showing that it does not need the highest image frequencies to make a correct decision.
The CNN-based detector, on the other hand, achieves a similar level only at $B$=0.3, relying largely on the 0.2-0.3 band.
Therefore, they appear to focus on different portions of the frequency spectrum, thus confirming the complementarity conjecture.

\section{Discussion}

We have studied the adversarial robustness of CLIP-based detectors for AI-generated images, also in comparison to popular CNN-based detectors.
Analysis of the numerical results and visual inspection of the spectra of adversarial patterns allow us to draw some significant lessons, 
summarized below.

CLIP-based detectors rely mainly on low image frequencies, in contrast to CNN-based ones that rely more on medium-high frequencies.
Although our observations are specific to media forensics, it is worth noting that similar results emerged in other computer vision fields as well \cite{benz2021adversarial, bai2022improving}.
The connection with frequency-based artifacts is especially important, since it clarifies the reason for which it is difficult to generalize between CNN and CLIP models.

CLIP-based detectors are not more robust to adversarial attacks than CNN-based detectors; however, we observed very limited transferability, especially between detectors using significantly different architectures, as CNN and Transformer backbones. This finding is also in-line with recent work in image classification \cite{mahmood2021robustness}.
The properties of adversarial pattern also depend on architectural and procedural details, but are rather stable for CLIP-based detectors. Finally,
these ones are more robust than CNNs to fake-to-real attacks, which are probably more relevant in a realistic scenario. 

Studying robustness to adversarial perturbations is important in itself for many real-world applications.
However, this study also sheds light on how detectors work and is therefore a valuable tool for developing more robust detectors.
Along these lines, in future work we will study the impact of adversarial attacks on generic forensic traces in synthetic images.

\section*{Acknowledgment}
We gratefully acknowledge the support of this research by a TUM-IAS Hans Fischer Senior Fellowship and a Google Gift. In addition, this work has received funding by the European Union under the Horizon Europe vera.ai project, Grant Agreement number 101070093, and was partially supported by SERICS (PE00000014) under the MUR National Recovery and Resilience Plan, funded by the European Union - NextGenerationEU. Finally, we thank the partnership and collaboration with the São Paulo Research Foundation (Fapesp) Horus project, Grant \#2023/12865-8.

\bibliographystyle{IEEEtran}
\bibliography{main}

\end{document}